\documentclass{article}

\usepackage[preprint]{neurips_2021}

\usepackage[utf8]{inputenc} 
\usepackage[T1]{fontenc}    
\usepackage{hyperref}       
\usepackage{url}            
\usepackage{booktabs}       
\usepackage{amsfonts}       
\usepackage{nicefrac}       
\usepackage{microtype}      
\usepackage{xcolor}         

\title{Using Psuedolabels for training Sentiment Classifiers makes the model generalize better across datasets}

\author{Natesh Reddy \\
  Paralleldots, Inc.  \\\And
  Muktabh Mayank Srivastava \\
  Paralleldots, Inc. \\\\\And
  \texttt{\{natesh,muktabh\}@paralleldots.com}
  }

\begin{document}

\maketitle

\begin{abstract}
    The problem statement addressed in this work is : For a public sentiment classification API, how can we set up a classifier that works well on different types of data, having limited ability to annotate data from across domains. We show that given a large amount of unannotated data from across different domains and pseudolabels on this dataset generated by a classifier trained on a small annotated dataset from one domain, we can train a sentiment classifier that generalizes better across different datasets.
\end{abstract}

\section{Introduction}
Sentiment Classification is the process of classifying a given text sentence as having positive, negative or neutral opinion. For example sentiment classification can be used for determining whether a movie or product review is positive / negative or neutral. In recently published works, Deep Learning based approaches have been shown to perform very well on sentiment classification datasets. These approaches typically work on a train-test set combination taken from the common domain. However, when a sentiment classification model is being made available as a public API (\href{https://github.com/ParallelDots/ParallelDots-Python-API}{API}) , the trainset data can be very different from the test data the algorithm will be used on by different users. In this work, we try to address the problem of training a common Sentiment Classifier which works well across multiple datasets coming from different domains. We start with training a teacher Sentiment Classifier on an initial dataset of annotated tweets and checking how it works on classification datasets from different domains. We then take a dataset of unannotated text documents [from across multiple domains] and generate sentiment psuedolabels for the documents using inferences of the teacher classifier trained in the first step. We then use the new dataset of pseudolabels to train a student sentiment classifier which works better on all classification datasets from different domains where the initial model was tested on.

\section{Related Work}
Sentiment Analysis is a widely studied problem in the field of NLP. There are different types of Sentiment Analysis datasets to evaluate models, some having just two polarities [Positive/Negative], others having three [Positive/Negative/Neutral] and there are datasets with more granularity as well. Two polarity benchmarks include : \href{https://www.kaggle.com/yelp-dataset/yelp-dataset}{Yelp binary classification} , \href{https://www.kaggle.com/lakshmi25npathi/imdb-dataset-of-50k-movie-reviews}{IMDB dataset} etc. \href{https://nlp.stanford.edu/sentiment/}{SST-5}   has 5 polarities as annotations. As a public API, we are supposed to give 3-way sentiment classification [Positive/Negative/Neutral] and this work focuses on such datasets only. The most famous public dataset example of this type is the \href{https://www.kaggle.com/kazanova/sentiment140}{Sentiment-140 dataset}, which is one of the datasets our models are evaluated on. 

Deep Learning models have recently shown to work very well on sentiment analysis problems. \citet{jiang-etal-2020-smart}, \citet{NEURIPS2020_44feb009}, \citet{NEURIPS2019_dc6a7e65} these models are typically language models finetuned in one way or another to train on classification tasks. 

Distant supervision using emojis has been a way to train sentiment models successfully. (\citet{Sentiment140}) This involves taking emojis like :-), :-D and so on as proxy for positive sentiment, :-( and emojis for other negative emotions as a proxy for negative sentiment and so on . In this work we finetune a Deep Learning model trained to predict emojis for Sentiment classification. (\citet{Felbo_2017})

Pseudolabels have been used to train state of the art models in Computer Vision. Such models work by training a teacher model on an annotated dataset and use inference by teacher models to create pseudolabels on an unannotated dataset. The psuedolabels are then used to train a noisy student model that works better than the teacher (\citet{xie2020selftraining}, \citet{zoph2020rethinking}). We show that training on Pseudolabels increases accuracy of Sentiment Classification model on datasets from different domains.

\section{Datasets}
We give a description of different datasets used in this work. The initial dataset used to train the teacher Sentiment Classifier is a dataset of 11000 annotated tweets, we call this dataset as TEACHER-TRAIN. The test set for this classifier of 1000 tweets is called TEACHER-TEST. We have 20 Million unannotated documents from different domains, for which the teacher Sentiment Classification model is used to create pseudolabels. This dataset of unannotated documents and their corresponding pseudolabels is called STUDENT-PSEUDO. Both teacher and student model are evaluated on datasets taken from different domains called evaluation datasets. Sentiment-140 dataset is a public dataset with Positive, Neutral and Negative sentiment annotations. \href{https://www.kaggle.com/kazanova/sentiment140}{Sentiment140} is used for evaluation. The other evaluation datasets used are not public datasets and are described in the following paragraph.

NEWS2016 is a dataset of 15000 annotated news article headings, each with a Positive, Negative or Neutral label. MARKETRESEARCH dataset is annotated data of 12300 annotated MARKETRESEARCH responses, again bearing one out of Positive, Negative or Neutral label. TWEETS25K is a dataset of 25000 tweets annotated with Positive, Negative or Neutral sentiment labels. 

Please note evaluation datasets are just used to measure accuracy of classifiers and are not involved in any training.

\section{Method}
We finetune a \href{https://deepmoji.mit.edu/}{DeepMoji} model to predict sentiment. The model uses an embedding layer of 256 dimensions to project each word into a vector space. A hyperbolic tangent activation function is used to enforce a constraint of each embedding dimension being within [-1, 1]. Two bidirectional LSTM layers with 1024 hidden units in each(512 in each direction) are used to capture the context of each word. Finally, an attention layer uses skip-connections to take all of these layers as input. 

We use the ‘chain-thaw’ approach to transfer learning, that sequentially unfreezes and fine-tunes a single layer at a time. This approach increases accuracy on the target task at the expense of extra computational power needed for the fine-tuning. By training each layer separately the model is able to adjust the individual patterns across the network with a reduced risk of overfitting. The sequential fine-tuning seems to have a regularizing effect similar to what has been examined with layer-wise training in the context of unsupervised learning.

More specifically, the chain-thaw approach first fine-tunes any new layers (often only a Softmax layer) to the target task until convergence on a validation set. Then the approach fine-tunes each layer individually starting from the first layer in the network. Lastly, the entire model is trained with all layers. Each time the model converges as measured on the validation set, the weights are reloaded to the best setting, thereby preventing overfitting in a similar manner to early stopping.

The first step is to finetune the DeepMoji model on the TEACHER-TRAIN dataset. The model created after this step is the teacher model. The teacher model is then run on the unannotated dataset to create the STUDENT-PSEUDO dataset.

Model Architecture - 

TorchMoji(

\hspace{10mm} (embed\_dropout): Dropout2d(p=0.1, inplace=False)
  
\hspace{10mm}  (lstm\_0): LSTMHardSigmoid(256, 512, batch\_first=True, bidirectional=True)
  
\hspace{10mm}  (lstm\_1): LSTMHardSigmoid(1024, 512, batch\_first=True, bidirectional=True)
  
\hspace{10mm}  (attention\_layer): Attention(2304, return attention=False)
  
\hspace{10mm}  (final\_dropout): Dropout(p=0.5, inplace=False)
  
\hspace{10mm}  (output\_layer): Sequential(
    
\hspace{12mm}    (0): Linear(in\_features=2304, out\_features=3, bias=True)
  
\hspace{10mm}  )

)

There are two ways in which we perform the step second step training :
\begin{enumerate}
  \item Teacher Fine Tuned : Where the teacher model is further finetuned on the pseudolabels of the STUDENT-PSEUDO dataset.
  \item Independent noisy student : A different DeepMoji backbone model is finetuned on combined STUDENT-PSEUDO and TEACHER-TRAIN dataset.
\end{enumerate}

The teacher model, Teacher Fine Tuned model and Independent noisy student are now run on various evaluation datasets and their accuracy is calculated. 

\section{Results}
The results of our experiment can be found in Table 1. We compare accuracy of different models on all evaluation datasets mentioned.

Looking at the results, it's very clear that the Independent Noisy student model works substantially better than the teacher model on every evaluation dataset. The only place where the teacher model does better is on the TEACHER-TEST dataset , which is the testset corresponding to TEACHER-TRAIN dataset, on which the teacher model was trained itself. This might be because TEACHER-TRAIN and TEACHER-TEST are sampled from the same domain and thus 20 Million pseudolabel based training forces the Independent Noisy Student model to forget dataset specific quirks.
\begin{table}\centering
  \caption{Results on various Datasets}
  \label{sample-table}
  \centering
  \begin{tabular}{p{3.5cm} p{1.8cm} p{3.2cm} p{3.2cm}}
    \toprule
    \cmidrule(l){1-2}
    \textbf{Dataset Name}     & \textbf{Accuracy of Teacher}     & \textbf{Accuracy of Teacher Finetuning}     & \textbf{Accuracy of Independent Noisy Student} \\
    \midrule
    \textbf{Sentiment-140}  & 68.87\%  & 64.05\%  & 73.69\%     \\
    \textbf{NEWS2016}  & 64.52\%  & 63.65\%  & 69.21\%     \\
    \textbf{MAKETRESEARCH}  & 82.29\%  & 83.09\%  & 90.09\%     \\
    \textbf{TWEETS25K}  & 76.87\%  & 74.86\%  & 77.52\%     \\
    \textbf{TEACHER-TEST}  & 75.8\%  & 75.7\%  & 74.46\%     \\
    \bottomrule
  \end{tabular}
\end{table}

\section{Conclusion}
We show in this work that training on pseudolabels generated by a teacher model can help a student model get better accuracy on datasets from different domains.


\newpage


\bibliography{neurips_2021}
\bibliographystyle{neurips_natbib}
\end{document}